\documentclass[10pt,twocolumn,letterpaper]{article}

\usepackage[pagenumbers]{cvpr} % To force page numbers, e.g. for an arXiv version
\definecolor{cvprblue}{rgb}{0.21,0.49,0.74}
\usepackage[pagebackref,breaklinks,colorlinks,allcolors=cvprblue]{hyperref}
\usepackage[accsupp]{axessibility}
\usepackage{mathtools}
\usepackage[normalem]{ulem}
\usepackage{amsmath}
\newcommand{\stkout}[1]{\ifmmode\text{\sout{\ensuremath{#1}}}\else\sout{#1}\fi}

\def\paperID{13768} % *** Enter the Paper ID here
\def\confName{CVPR}
\def\confYear{2025}

\begin{document}

\title{ERUPT: Efficient Rendering with Unposed Patch Transformer }

\author{
Maxim V. Shugaev\footnotemark[1] \qquad Vincent Chen\footnotemark[1]\qquad Maxim Karrenbach\footnotemark[2] \\
Kyle Ashley\footnotemark[1] \qquad Bridget Kennedy\footnotemark[1] \qquad Naresh P. Cuntoor\footnotemark[1] \\
\textsuperscript{*}BlueHalo \qquad \textsuperscript{\textdagger}Carnegie Mellon University \\
{\tt\small maxim.shugaev@bluehalo.com}\\
% \and
}
\maketitle

%%%%%%%%% ABSTRACT
\begin{abstract}
This work addresses the problem of novel view synthesis in diverse scenes from small collections of RGB images. We propose ERUPT (Efficient Rendering with Unposed Patch Transformer) a state-of-the-art scene reconstruction model capable of efficient scene rendering using unposed imagery. We introduce patch-based querying, in contrast to existing pixel-based queries, to reduce the compute required to render a target view. This makes our model highly efficient both during training and at inference, capable of rendering at 600 fps on commercial hardware. Notably, our model is designed to use a learned latent camera pose which allows for training using unposed targets in datasets with sparse or inaccurate ground truth camera pose. We show that our approach can generalize on large real-world data and introduce a new benchmark dataset (MSVS-1M) for latent view synthesis using street-view imagery collected from Mapillary. In contrast to NeRF and Gaussian Splatting, which require dense imagery and precise metadata, ERUPT can render novel views of arbitrary scenes with as few as five unposed input images. ERUPT achieves better rendered image quality than current state-of-the-art methods for unposed image synthesis tasks, reduces labeled data requirements by ~95\% and decreases computational requirements by an order of magnitude, providing efficient novel view synthesis for diverse real-world scenes. 
\end{abstract}

\section{Introduction}

Recovering 3D scene representation from a few images is a fundamental challenge in computer vision with wide-reaching applications in navigation, robotics, and augmented reality. Recent advances in learning implicit (\textit{e.g.} NeRF \cite{Mildenhall2022}) and explicit (\textit{e.g.} Gaussian Splatting \cite{Kerbl2023}) scene representation have shown impressive capability in rendering high-fidelity, novel views of complex environments. These models typically rely on dense imagery with precise camera pose to reconstruct highly accurate models which can render novel views of the scene. However, one problem, which is central to both the NeRF and Gaussian Splatting frameworks, is that a new network or a set of Gaussians must be trained for each scene. This requires significant training time and limits the model’s ability to incorporate prior knowledge in cases such as limited input views or large camera shifts. Several methods have been introduced to reduce this scene dependency, by including depth \cite{Deng2022}, normal \cite{Turkulainen2024}, semantic \cite{Bao2023} and other priors to use information from external data as a basis for learning a particular scene. Yet the problem remains that these frameworks are generally unable to perform novel view synthesis without training on a sufficiently large number of input views and associated camera information for each new scene.

More recent work has explored techniques for generalized 3D reconstruction with limited imagery using a latent scene representations transformer (SRT) \cite{Sajjadi2022} and learnable pose estimator (RUST) \cite{Sajjadi2023} to reduce reliance on precise camera poses. By training on large and diverse datasets these models learn to reconstruct arbitrary scenes using fewer images and limited camera pose information.

While SRT allows images to be queried directly using an arbitrary camera pose, it does not provide a mechanism for learning from unposed target views. This makes it difficult to apply in datasets where ground truth camera pose is noisy or unknown. RUST addresses this limitation using a learnable pose estimator but requires a portion of the target image to query the model, which limits control of the camera at inference. Both methods require significant compute during training and inference using individual queries to the decoder for each pixel in the rendered images. Furthermore, existing benchmark datasets of sufficient scale are either composed of synthetic data (MSN) or are proprietary datasets not available to the public, making verification and real-world application of these techniques difficult.

To address these limitations we propose ERUPT, a computationally effective approach for novel view synthesis trained on posed and unposed target images, providing direct camera control at inference. Our model includes a learnable latent camera pose which is sampled in conjunction with sparse ground truth pose to address cases of sparse or inaccurate target camera pose. We introduce a novel decoding strategy using patch-based rays instead of pixel-based rays to improve computation efficiency of rendering by an order of magnitude over existing models. To accelerate convergence, during training we generate multiple targets reusing scene representation and apply a pretrained feature extractor for transfer learning. We compare our model with existing methods using the MSN benchmark dataset, achieving state-of-the-art performance even compared to methods which rely on 100\% posed target images. Additionally, we introduce a new real-world view synthesis dataset called MSVS-1M (Mapillary Street-view Synthesis 1-million) to allow future studies to conduct direct comparisons on real-world data. 

\noindent In summary our contributions include: 
\begin{enumerate}
  \item An efficient network architecture and training regime for latent-based novel view synthesis trained on posed and unposed target images which provides direct camera control at inference. We show that training on data with only 5\% posed targets results in minimal performance degradation compared to the baseline with target poses provided.
  \item Introduction of patch-based rays in the decoder instead of pixel-based rays results in an order of magnitude reduction in rendering and training time as well as VRAM requirements. Specifically, the synthesis of 224×224 images can be performed at a rate of over 600 fps.
  \item An optimized model architecture for image synthesis with unknown input pose. The produced models establish a new SOTA for image synthesis on MSN dataset with unknown input pose in terms of PSNR and SSIM.
  \item Using ERUPT as a foundation, we investigate the use of adversarial learning and diffusion-based rendering for latent novel view synthesis. Both approaches drastically improve the perceptual quality of the output and outperform previous perception based SOTA on MSN dataset.
  \item A new multi-scene view synthesis dataset, MSVS-1M, curated from Mapillary imagery to facilitate large-scale experiments in generalized real-world view synthesis.
\end{enumerate}

% The scope of the current work is unknown input and known camera: relevant from the practical point of view for novel view synthesis. Our main focus is creation of a computationally efficient setup. 

% \noindent In contrast to NERF and Gaussian Splatting, which require fitting a new model for each scene and take many hours to produce the output, for latent synthesis models consideration of a completely novel scene may take a sub second time (with 100s fps rendering when the scene representation is produced). After training on a large corpus of scenes, these models, in addition, reuse prior knowledge to generalize well for large view shifts and completely novel views in comparison to NERF and Gaussian Splatting type of approaches, which are mainly bounded to the input provided with the current set of images. Following recent SRT \cite{Sajjadi2022} and RUST \cite{Sajjadi2023} work on latent image synthesis we have designed and developed ERUPT (Efficient Rendering with Unposed Patch Transformer). Since large scale (millions of scenes) outdoor scene datasets with precise camera parameters do not exist, we, similar to RUST approach, focus on scene synthesis with unknown camera parameters of input images.  

\begin{figure*}
\centering
\includegraphics[width=1.0\linewidth]{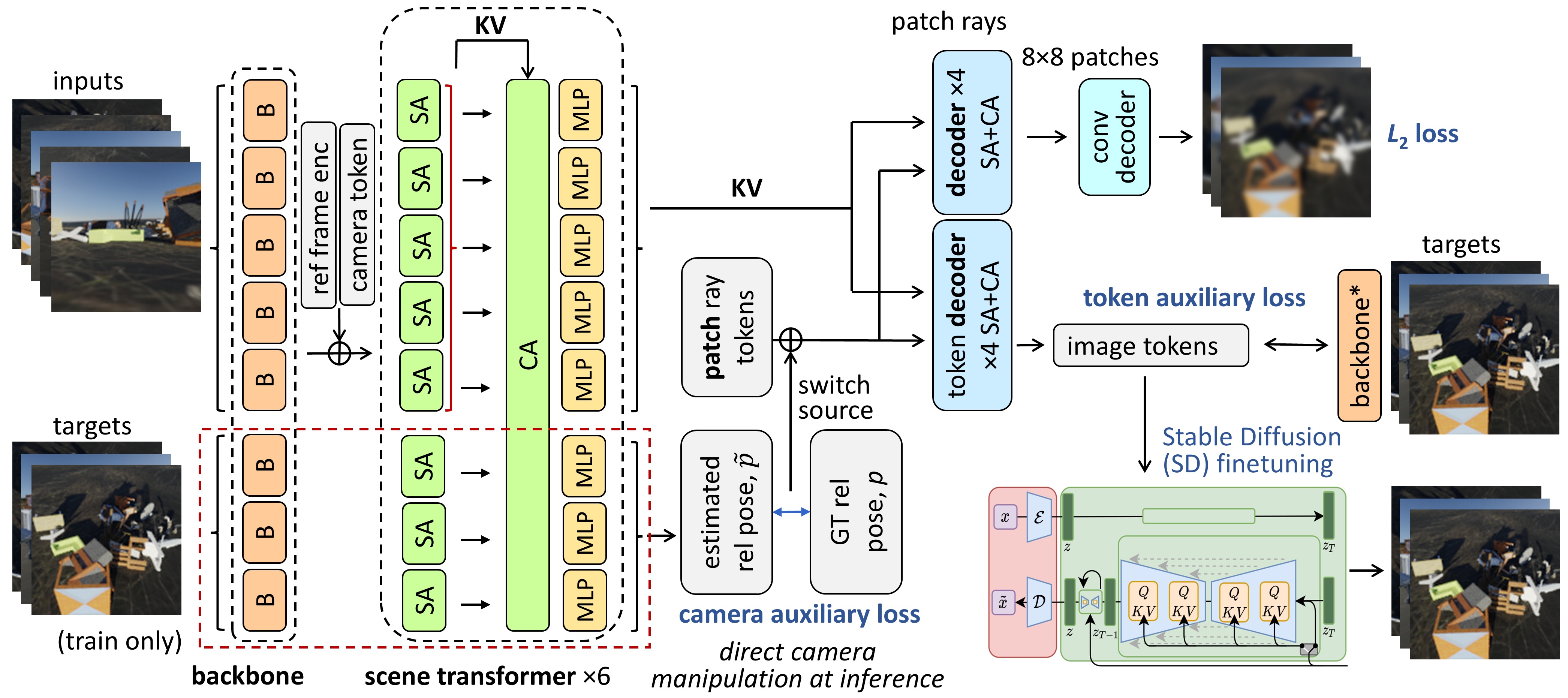}
\caption{\label{fig:EruptSchematic}Schematic illustration of ERUPT model. B refers to backbone (feature extractor), while SA and CA refer to self- and cross- attention.}
\end{figure*}

\section{Related Work}

\subsection{NeRF and Gaussian Splatting}

In the realm of novel view synthesis, two main frameworks have stood out at the forefront. Neural Radiance Fields (NeRF) \cite{Mildenhall2022} has approached the problem by predicting the color and density of points in a 3D representation space through neural networks as an implicit representation. By modifying the neural network and training scheme, this framework can serve as a base for further works, like object editing/removal \cite{Kundu2022}, appearance or lighting editing \cite{MartinBrualla2021}, or dynamic scene synthesis \cite{Attal2021}. More recently, 3D Gaussian Splatting \cite{Kerbl2023} has been proposed and shown to reduce the high rendering time cost of NeRFs at the cost of requiring most detailed hyperparameters. The 3DGS framework uses an explicit representation of the 3D scene based on 3D Gaussians with spherical harmonics defining the color and opacity. The low rendering time is a draw to this type of framework for inference-heavy applications, and the granularity of the 3D Gaussians allow for techniques such as per-Gaussian editing \cite{Bae2024}, Gaussian grouping \cite{Ye2024}, or appearance conditioning \cite{Dahmani2024}.

Both frameworks have been built upon by numerous other works to reduce limitations including reduced data dependency, improved rendering quality and optimizations for rendering speed. For example, frameworks like Sparse-GS \cite{xiong2023sparsegs} and SparseNeRF \cite{Wang2023} have been developed to address performance degradation due to limited number of training views by using alternate strategies like depth information. To address the problems of large training times, both frameworks have implemented techniques, like the NeRF-based multi-resolution hash encodings in Instant-NGP \cite{Muller2022} or 3DGS-based vector quantization \cite{Navaneet2024} in Compact3D. These frameworks have even been combined to get the best of both techniques via hybrid models \cite{Malarz2024} called Viewing Direction Gaussian Splatting, which uses the Gaussian distributions and includes a neural network which interprets the Gaussian parameters and viewing directions. 

However, one problem which is central to both the NeRF and 3DGS frameworks is that a new network or a new set of Gaussians must be trained and fitted for each scene, requiring training time. Works have indeed been done to reduce this scene dependency, by including depth \cite{Deng2022}, normal \cite{Turkulainen2024}, semantic \cite{Bao2023} and other priors to use information from external data as a basis for learning a particular scene. Yet the problem remains in that these frameworks are generally unable to perform novel view synthesis without training on numerous input views with precise metadata provided, which results in a necessary time to train on a novel scene. 

\subsection{Posed and unposed SRT}
Latent scene representation approaches mitigate limitations of NeRF and Gaussian Splatting on transferring prior knowledge by considering a generalized encoder-decoder architecture. This model, after training on a large corpus of scenes, creates a scene representation based on the provided input images and allows rendering novel views without the need to retrain the model for the particular scene. One of the early works that has demonstrated the scalability of the latent scene representation approach to complex scenes is the Scene Representation Transformer (SRT) \cite{Sajjadi2022}. The proposed model consists of a light CNN applied to input images and the corresponding input poses and creates a set of tokens. These tokens are then passed to a scene representation transformer. The transformer performs mixing of all input views and creates a latent scene representation. Finally, this representation is queried in a set of decoder blocks based on rays corresponding to each target pixel. In \cite{jin2024lvsmlargeviewsynthesis} the architecture was extended to decoder only. In several studies \cite{yu2021pixelnerfneuralradiancefields, wang2021ibrnetlearningmultiviewimagebased} the NeRF framework is extended with learnable mechanism to enable novel view synthesis based on a few or even single input image. However, they have limitations or under-perform SRT \cite{Sajjadi2022}. The SRT model requires knowledge of the camera parameters for all input and target images at training and inference. This limitation is partially addressed in UpSRT that uses known pose only for the first input image and the target. However, as illustrated in \cite{Sajjadi2023}, training on perturbed target poses (which corresponds to the realistic setup of using estimated target poses) results in significant performance degradation.

\subsection{RUST}
To address the limitation of SRT work on training on scenes with unknown target poses, Latent Neural Scene Representations from Unposed Imagery (RUST) \cite{Sajjadi2023} model was proposed. In this model, the UpSRT setup is supplemented with an additional pose estimation module that takes half of the target image and the latent scene representation to compute a latent pose (8-dimensional vector), which is used to query the scene representation in the decoder (instead of the real target pose) and generate the remaining half of the target image. This setup enables robust training on completely unposed imagery sets. Generation of the particular target view at inference still requires half of the target image, which limits the views that can be generated. Although, \cite{Sajjadi2023} demonstrates the capability of RUST to perform shifts and rotations of the camera by manipulating principal components of the latent pose vector, precise camera control is challenging and may involve manual tuning steps.

\subsection{DORSal} 
Diffusion for Object-Centric Representation of Scenes (DORSal) \cite{jabri2024dorsaldiffusionobjectcentricrepresentations}  is a recent work that considers a diffusion decoder to perform scene rendering based on the scene representation created with a frozen Object Scene Representation Transformer (OSRT) \cite{sajjadi2022objectscenerepresentationtransformer}. Use of the diffusion model drastically improves the quality of the generated scenes improving FID by 5-10×.

\section{Methodology}

\subsection{Model}

\cref{fig:EruptSchematic} schematically illustrates the ERUPT model developed in our work. The input scene consists of an unordered set of (unposed) views, which are processed with a transformer-based feature extractor. A reference frame encoding and positional encoding are added to the collection of image features, and the resulting set of tokens is expanded with a learnable camera token. Next, the token sequence is processed with a scene transformer designed to generate compact scene and camera representations. This transformer alternates token mixing within each image and between images within the entire scene, producing the scene representation and camera pose estimation for each image (which are computed with respect to the reference frame). The resulting scene representation is queried for each patch-ray in the image decoder and token decoder, which produce the image output and a sequence of tokens matching the target image. Using patch rays instead of pixel rays drastically improves the computational efficiency and enables the use of the generated sequence of tokens to condition a fine-tuned Stable Diffusion (SD) model \cite{Rombach2022}. 

At the training stage, a set of target images is also passed through the feature extractor and scene transformer to estimate the target camera parameters. Then the query camera parameters are selected from the estimated and the ground truth poses to maintain the RUST advantage of training on unposed images or images with inaccurate camera poses and at the same time to have the ability of direct camera manipulation at inference. Below we discuss each part of the model in details.

\textbf{Scene transformer}: The objective of the scene transformer is to extract a compact representation of the scene from a small set of images which is flexible enough to function on diverse scenes. In the SRT setup proposed in \cite{Sajjadi2022}, camera information is attached to each image which enables effective discrimination of patches from different input views and scene understanding. However, this model architecture cannot be easily adapted to the case of unknown camera parameters of input images. The approach proposed in RUST \cite{Sajjadi2023} applies a similar transformer setup to SRT except, instead of encoding the camera parameters, a simple camera encoding is used to distinguish the reference frame from the remaining input images, \textit{i.e.} all non-reference images are not distinguishable in the model, and scene understanding is limited. To address this fundamental problem we have redesigned the architecture of the scene transformer. Specifically, we alternate mixing of tokens within images (self-attention between image tokens) with mixing of tokens between images for the entire scene (cross-attention of image tokens and all scene tokens), \textit{i.e.} the model while exchanging the information between input views has a capability to distinguish tokens from different images and build the camera representation for each input even if the camera parameters are unknown. This design alleviates the limitations of simple token mixing used by existing approaches to build a robust scene representation for unposed inputs. The scene transformer consists of 6 transformer blocks in all experiments.

\textbf{Pose estimation}: Our scene transformer network enables direct camera parameter estimation for each image without including a separate model branch (as used in RUST). We add a learnable camera token to each image to aggregate the camera information and the relative camera pose of each image to the reference. An auxiliary loss is applied on the learnable camera pose that guides the camera estimation and may improve the scene understanding. In our setup we use $L_{2}$ loss applied to the relative camera position and negative cosine loss applied to 3D vectors describing the relative camera view. To avoid information leakage from the target images, the cross-attention part of the scene transformer uses the combination of tokens from input images only as the key-value (KV) input.

\textbf{Target camera switching}: A major contribution of our method is the ability to query images from the model in a fully-controllable manner using arbitrary target poses. In contrast, RUST uses pose estimation on half of the target image followed by an estimated latent pose to render the remaining part. While this enables training and inference on input images with unknown camera pose, it requires a portion of the target view to be available at inference time which limits the model to only views which have been partially observed. This issue is discussed in the RUST paper in detail, and the use of several principal components of the latent pose is suggested to describe a particular motion of the camera. However, precise camera manipulation is challenging and may involve manual tuning steps for each model.

We introduce the ability to manipulate the target camera at inference time while maintaining the advantage of training on unposed images. To accomplish this, during training, we concatenate the estimated latent pose with sine encoded scaled camera parameters \cite{Vaswani2017,Sajjadi2022} and randomly sample between three modes (1) providing only the estimated latent camera pose, (2) providing only the ground truth camera pose and (3) providing both estimated and ground truth camera pose. These modes are sampled evenly with 1/3 probability during training, unless otherwise specified. At inference we use only the encoded target image pose, which enables direct camera control. Meanwhile, use of the estimated latent camera pose pathway may be helpful for training on images without known pose or with inaccurate camera pose estimation: when the target view contradicts the provided pose, the model may use the latent pose pathway to generate the expected output and enable robust training. We conducted an ablation study to assess the robustness of the approach when highly limited ground truth is available and found that using just 5\% ground truth poses resulted in minimal degradation (\cref{tab:ablation-result}).

\textbf{Patch based decoder}: Existing latent scene representation models require significant computation for both training and inference. SRT and RUST use pixel-based decoding, \textit{i.e.} querying the scene representation for each pixel, which introduces an enormous overhead both in terms of the speed and VRAM consumption, as discussed in the section on the computational performance. To improve the computational performance we use patch base rays with patch size of 8×8 for the image decoder and patch size matching the backbone for the token decoder. Using this method ERUPT recovers 64 pixels per query instead of 1 pixel as done in pixel-based decoding models.

The patch ray queries are learned during training and initialized with sine positional encoding \cite{Vaswani2017}. The target camera representation is concatenated to each query token. Each decoder is composed of 4 standard transformer decoder blocks \cite{Vaswani2017} alternating self-attention within ray queries and cross-attention with the scene representation. The image decoder is followed with 3 convolutional pixel unshuffle \cite{Shi2016} up-sampling blocks producing the final image. $L_{2}$ loss is applied to the pixel output. In addition to the image decoder, we use token decoder that is trained to match the embedding produced by the pretrained backbone on the target image, \textit{i.e.} enforce the semantics of the generated output. ArcGeo auxiliary loss \cite{Shugaev2024} is applied to each token of the produced embedding within the target image.

\textbf{Feature extractor}: Both SRT and RUST use a simple convolutional encoder trained together with the rest of the model to extract image tokens, while in \cite{Venkat2023} ImageNet initialized ResNet18 is applied. We follow the idea of using a pretrained model to extract reliable features and accelerate training. In our setup we upscale input images to 224 size and apply DINOv2 \cite{Oquab2024} as a feature extractor. We use DINOv2 with additional registers, which is demonstrated to have better performance and feature map interpretability \cite{Darcet2024}. This model is suggested to have a robust single image 3D understanding as well as multi-view consistency \cite{10656175}, which may be helpful for 3d scene synthesis. The produced feature map has 16×16 dimensions (similar to RUST) plus 1+4 extra classification and register tokens. Learnable positional encoding is added to the produced sequence of tokens before feeding to the scene transformer together with the reference frame encoding. We freeze the backbone in training and perform an additional set of experiments with enabling LORA \cite{Hu2022} fine-tuning of the backbone after the initial model training. The computational cost of the forward pass for DINOv2 B model is comparable to the forward and backward pass cost for the extractor used in SRT and RUST, and use of the pretrained model introduces a minimum overhead during training.

\textbf{GAN setup}: In several experiments we performed fine-tuning the models to improve the perceptual quality of the output. Generative adversarial networks (GANs) has been successfully leveraged in related applications of image generation \cite{Liang2021, ledig2017photorealisticsingleimagesuperresolution} to improve realism and align distributions of synthesized imagery. We explored their impact on latent view synthesis in ERUPT, and follow existing adversarial formulations \cite{Liang2021}. We replaced $L_{2}$ image output loss with a combination of $L_{1}$, perception, and GAN losses. We use the SwinIR \cite{Liang2021} configuration from \textit{KAIR} repository \cite{Zhang2019} as a starting point. 

\textbf{Stable Diffusion based rendering}: 
ERUPT’s patch-based decoder not only improves computational efficiency but also enables the use of pretrained generative models, \textit{e.g.} Stable Diffusion (SD) \cite{Rombach2022}, as a rendering tool. Specifically, in a series of experiments we applied SD 2.1 from the \textit{Diffusers} repository \cite{vonPlaten2022} to perform the final step of image generation instead of image decoder. In this setup we freeze the ERUPT model and use the token decoder output with an additional positional encoding as a prompt to the diffusion model. Then we perform a full fine-tuning of SD U-Net for 20 epochs on the considered dataset with rescaling the target images to 512×512. SD model is quite flexible to prompting adjustment, and relatively short fine-tuning is sufficient to enable generation even based on non-CLIP tokens.

\textbf{Model Size}: We consider two ERUPT models sizes. Our ERUPT-B model uses embedding dimension of 768 and contains a comparable number of trainable parameters to the 74M SRT model \cite{Sajjadi2022}. Specifically, the scene transformer and image decoder contain 57.4M and 15M parameters, respectively (72.4M in total). In addition, ERUPT-B contains a 85.7M parameter frozen DINOv2-B feature extractor and a 38.9M parameter token decoder, which may be omitted at the inference with image decoder. ERUPT-L uses embedding dimension of 1024 and contains 102M and 23.9M scene transformer and image decoder (125.9M in total). In addition, this model uses 194.9M parameter frozen DINOv2-L extractor and 68.9M parameter token decoder.

\subsection{Data}

\textit{MSN dataset} 
(MSN) is a novel view synthesis dataset introduced in \cite{Sajjadi2022} containing 1M train and 100k test scenes. Each scene consists of 10 128×128 images representing viewpoints selected by uniformly sampling within a half-sphere shell. This dataset is created with photorealistic ray tracing, and each scene has 16-31 ShapeNet objects dropped at random locations within an invisible bounding box. The dataset includes precise camera information. In the standard training/inference setup the model receives 5 randomly selected input views per scene, and the target is sampled from the remaining views.

\textit{MSVS-1M dataset} \cite{msvs1m} is a collection of real-world panoramic images captured by a vehicle along its trajectory \cite{Warburg_2020_CVPR}. This dataset aims to tackle real-world challenges not captured by synthetic MSN data such as noisy camera pose, lighting/semantic variability, and diverse environments. The data is similar to the private StreetView dataset used in \cite{Sajjadi2022, Sajjadi2023}, but is collected through publicly available sources and released to the community under Mapillary’s CC-BY-SA license. To create the dataset, we randomly selected 10 locations in the world and sampled street view panoramas within 60 kilometers of each location. We selected only images with corrected camera parameters obtained through Mapillary's structure-from-motion process \cite{opensfm} and performed filtering to reduce the fraction of blurry images and images with rain droplets. We chose image sequences with a distance of 0.5-10 m between image acquisition location and a total length of at least 6. The  number of filtered images is approximately 1M, corresponding to 32k sequences. At training/evaluation we perform dynamic scene generation with details provided in the Supplementary Materials.

\subsection{Training Setup}

We train our models with AdamW \cite{Loshchilov2019} optimizer for 160 epochs with the maximum learning rate of $5 \times 10^{-4}$ and cosine schedule (10k warmup steps). The batch size is 128, and weight decay is set to 0.05. We use 5 target images per scene (if other values are not specified) to reuse the scene representation and improve the efficiency of training, even if the time per epoch is approximately doubled compared to the single target setup. Our models exhibit competitive results while training is more than 5 times shorter than 1000 epochs used in SRT setup \cite{Sajjadi2022}.

In LORA fine-tuning \cite{Hu2022} we added $r$=32 adapters to $qkv$ and output projection in attention and to MLP layers of the frozen DINOv2 backbone, enabling adjustment of the model to the task of 3D scene reconstruction and to synthetic MSN input. LORA added 5M parameters to B and 13M parameters to L model. As the starting point we used a model trained with a frozen feature extractor and continued training for 32 additional epochs with LORA and the maximum learning rate of $2 \times 10^{-5}$.

In GAN experiments we merged LORA with the frozen backbone and trained ERUPT for an additional 64 epochs (with a single target image), adding perception and GAN losses while using a maximum learning rate of $10^{-5}$. In SD experiments we performed full fine-tuning of SD U-Net (while ERUPT is frozen) for 20 epochs with the maximum learning rate of $10^{-5}$.

\section{Experiments}

% \begin{table*}[h]
%     \centering
%     \begin{tabular}{@{\extracolsep{\fill}} l cccccc @{}}
%         \toprule
%         Setup & & PSNR↑ & SSIM↑ & LPIPS↓_{Alex} & LPIPS↓_{VGG} & FID↓ \\
%         \midrule
%         SRT & $p_x$, $p_y$& 23.41 & 0.697 & Value 3 & 0.369 & Value 5 \\
%         SRT &$p_x$, $p_y$ & 25.93 & Value 2 & Value 3 & 0.237 & 67.29 \\
%         UpSRT & $\stkout{p_x}$, $p_y$& 23.03 & 0.683 & 0.300 & 0.362 & Value 5 \\
%         RUST &$\stkout{p_x}$, $\stkout{p_y}$ & 23.49 & 0.703 & 0.287 & 0.351 & Value 5 \\
%         DORSal &$p_x$, $p_y$ & 18.99 & Value 2 & Value 3 & 0.265 & 9.00 \\
%         ERUPT B + LORA &$\stkout{p_x}$, $p_y$ & 24.69 & 0.7486 & 0.2690 & 0.3598 & 99.7 \\
%         ERUPT L + LORA &$\stkout{p_x}$, $p_y$ & 25.26 & 0.7692 & 0.2440 & 0.3397 & 91.1 \\
%         ERUPT B + GAN &$\stkout{p_x}$, $p_y$ & 23.38 & 0.7132 & 0.1175 & 0.2035 & 7.45 \\
%         ERUPT B + SD &$\stkout{p_x}$, $p_y$ & 21.06 & 0.6373 & 0.1256 & 0.2338 & 6.89 \\
%         \bottomrule
%     \end{tabular}
%     \caption{Comparison of ERUPT to latent scene representation methods on MSN dataset}
%     \label{tab:erupt-result}
% \end{table*}

\begin{table}[h]
    \centering
    \begin{tabular}{@{\extracolsep{\fill}} l ccccc @{}}
        \toprule
        Method & & PSNR↑ & SSIM↑ & LPIPS↓ & FID↓ \\
        \midrule
        SRT & \color{green}$p_x$ $p_y$& 23.41 & 0.697  & 0.369 & - \\
        SRT* & \color{green}$p_x$ $p_y$ & \textbf{25.93} & - & 0.237 & 67.29 \\
        UpSRT & \color{red}$\stkout{p_x}$ \color{green}$p_y$& 23.03 & 0.683& 0.362 & - \\
        RUST & \color{red}$\stkout{p_x}$ $\stkout{p_y}$ & 23.49 & 0.703 & 0.351 & - \\
        DORSal & \color{green}$p_x$ $p_y$ & 18.99 & - & 0.265 & 9.00 \\
        \midrule
        % ERUPT: & & & & & \\
        \multicolumn{2}{@{} l}{ERUPT (Ours):} & & & & \\
        \textit{B+LORA} & \color{red}$\stkout{p_x}$ \color{orange}$p_y$ & 24.69 & 0.749 & 0.360 & 99.7 \\
        \textit{L+LORA} & \color{red}$\stkout{p_x}$ \color{orange}$p_y$ & \textbf{25.26} & \textbf{0.769} & 0.340 & 91.1 \\
        \textit{B+GAN} & \color{red}$\stkout{p_x}$ \color{orange}$p_y$ & 23.38 & 0.713 & \textbf{0.204} & 7.45 \\
        \textit{B+SD} & \color{red}$\stkout{p_x}$ \color{orange}$p_y$ & 21.06 & 0.637 & 0.234 & \textbf{6.89} \\
        \bottomrule
    \end{tabular}
    \caption{Comparison of ERUPT to latent scene representation methods on MSN dataset. ${p_x}$,${p_y}$ refer to input and target poses. Colors: green - available pose, red - not available pose, orange (ERUPT) - sparse camera pose at training and available pose at inference. SRT metrics are reported in \cite{Sajjadi2022}; UpSRT and RUST in \cite{Sajjadi2023}; and SRT* and DORSal in \cite{jabri2024dorsaldiffusionobjectcentricrepresentations}.}
    \label{tab:erupt-result}
\end{table}

% \begin{table*}[h]
%     \centering
%     \begin{tabular}{@{\extracolsep{\fill}} l ccccc @{}}
%         \toprule
%         Setup & & PSNR↑ & SSIM↑ & LPIPS↓ & FID↓ \\
%         \midrule
%         SRT & \color{green}$p_x$, $p_y$& 23.41 & 0.697  & 0.369 & - \\
%         SRT & \color{green}$p_x$, $p_y$ & 25.93 & - & 0.237 & 67.29 \\
%         UpSRT & \color{red}$\stkout{p_x}$, \color{green}$p_y$& 23.03 & 0.683& 0.362 & - \\
%         RUST & \color{red}$\stkout{p_x}$, $\stkout{p_y}$ & 23.49 & 0.703 & 0.351 & - \\
%         DORSal & \color{green}$p_x$, $p_y$ & 18.99 & - & 0.265 & 9.00 \\
%         \midrule
%         ERUPT B + LORA & \color{red}$\stkout{p_x}$, \color{yellow}$p_y$ & 24.69 & 0.749 & 0.360 & 99.7 \\
%         ERUPT L + LORA & \color{red}$\stkout{p_x}$, \color{yellow}$p_y$ & 25.26 & 0.769 & 0.340 & 91.1 \\
%         ERUPT B + GAN & \color{red}$\stkout{p_x}$, \color{yellow}$p_y$ & 23.38 & 0.713 & 0.204 & 7.45 \\
%         ERUPT B + SD & \color{red}$\stkout{p_x}$, \color{yellow}$p_y$ & 21.06 & 0.637 & 0.234 & 6.89 \\
%         \bottomrule
%     \end{tabular}
%     \caption{Comparison of ERUPT to latent scene representation methods on MSN dataset. Colors: green - available pose, red - not available pose, orange (ERUPT) - sparse camera pose at training and available pose at inference. }
%     \label{tab:erupt-result}
% \end{table*}

\subsection{Novel view synthesis}
We compare ERUPT with several existing methods. SRT provides a reference for synthesizing novel views under ideal conditions with known ground truth camera parameters both for input and target images and uses a transformer-based encoder decoder model. The original SRT architecture is improved in the later publications \cite{sajjadi2022objectscenerepresentationtransformer, Sajjadi2023}. UpSRT and RUST are included in our evaluation to demonstrate results for cases with unknown input image pose and known/unknown target pose, while DORSal showcases diffusion-based model performance. Quantitative results are summarized in \cref{tab:erupt-result}.

\begin{figure}
\centering
\includegraphics[width=1.0\linewidth]{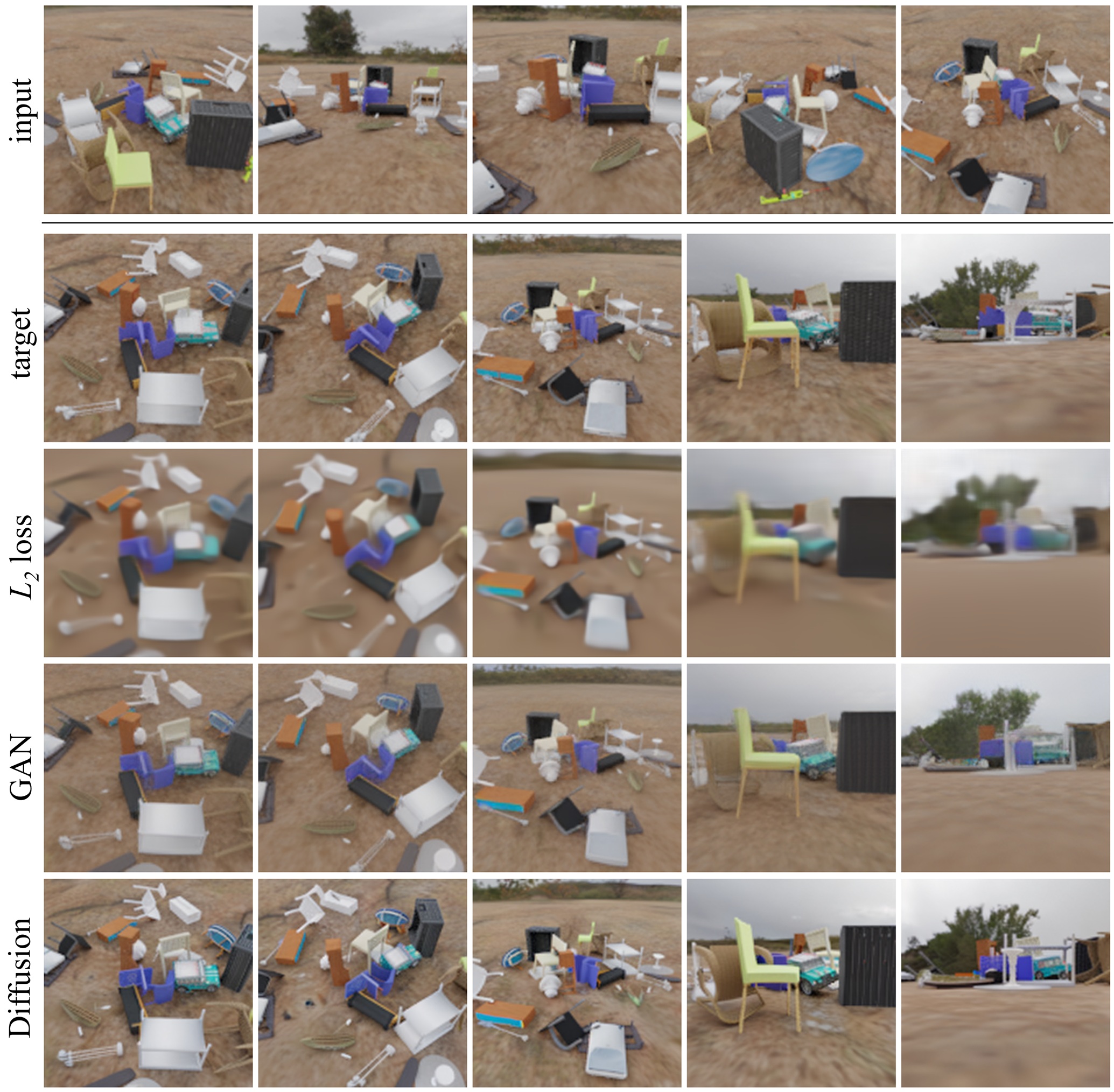}
\caption{\label{fig:EruptResults}Qualitative results on MSN: input scene (5 images), target views (5 images), ERUPT B + LORA, ERUPT B + GAN, ERUPT B + Diffusion.}
\end{figure}

ERUPT B and L with LORA finetuned feature extractor strongly outperforms UpSRT/RUST in terms of PSNR/SSIM and reaches the performance close to the optimized SRT despite of unknown input poses and mixture of known and unknown target poses used in training. The perceptual quality of the output is comparable to unoptimized SRT, UpSRT, and RUST, and the larger gap between PSNR and LPIPS may be attributed to use of patch-ray instead of pixel-ray based decoder. While our model trained with simple $L_{2}$ loss yields blurry images due to high uncertainty of the scene content, it can be significantly improved after a short fine-tuning in a GAN based setup, outperforming the perceptual quality reported in literature on MSN dataset. Both, GAN and diffusion setups outperform DORSal (the current SOTA on MSN) in terms of FID. Synthesized images generated using various ERUPT configurations are illustrated in \cref{fig:EruptResults}. Despite the superior visual quality of diffusion based output, it has only slightly better FID than the GAN setup, which may miss fine details. Finally, in contrast to DORSal, each output frame is synthesized independently on each other, which may result in inconsistency. Despite this limitation, the model is able to produce coherent scene output even if the particular item is presented in 1-2 input views, as a car in \cref{fig:EruptResults}. Further details on the model output are provided in the supplementary materials.

% \begin{table*}[h]
%     \centering
%     \begin{tabular}{lcccccc}
%         \toprule
%         Setup & PSNR↑ & SSIM↑ & LPIPS↓_{Alex} & LPIPS↓_{VGG} & FID↓ \\
%         \midrule
%         Baseline(b) & 23.85 & 0.7175 & 0.3020 & 0.3823 & 105 \\
%         Single target & 23.43 & 0.7001 & 0.3247 & 0.4008 & 115 \\
%         Patch RUST & 23.20 & 0.6903 & 0.3373 & 0.4108 & 119 \\
%         B model + LORA & 24.69 & 0.7486 & 0.2690 & 0.3598 & 99.7 \\
%         L model & 24.29 & 0.7349 & 0.2805 & 0.3672 & 96.4 \\
%         L model + LORA & 25.26 & 0.7692 & 0.2440 & 0.3397 & 91.1 \\
%         5\% known poses & 23.55 & 0.7060 & 0.3141 & 0.3922 & 108 \\
%         \bottomrule
%     \end{tabular}
%     \caption{Ablation Results on MSN dataset}
%     \label{tab:ablation-result}
% \end{table*}

\begin{table}[h]
    \centering
    \begin{tabular}{lcccccc}
        \toprule
        Method & PSNR↑ & SSIM↑ & LPIPS↓ & FID↓ \\
        \midrule
        \textit{baseline} & 23.85 & 0.718 & 0.382 & 105 \\
        Single target & 23.43 & 0.700 & 0.401 & 115 \\
        Patch RUST & 23.20 & 0.690 & 0.411 & 119 \\
        ERUPT B+LORA & 24.69 & 0.749 & 0.360 & 99.7 \\
        ERUPT L & 24.29 & 0.735 & 0.367 & 96.4 \\
        ERUPT L+LORA & 25.26 & 0.769 & 0.349 & 91.1 \\
        5\% known poses & 23.55 & 0.706 & 0.392 & 108 \\
        \bottomrule
    \end{tabular}
    \caption{Ablation results on MSN dataset. \textit{baseline} refers to ERUPT B.}
    \label{tab:ablation-result}
\end{table}

\subsection{Ablation Results}
We now show results from a series of ablation studies on the MSN dataset in \cref{tab:ablation-result}. As a baseline ERUPT-B setup is used. Reduction of the number of target images from 5 to 1 in training reduces PSNR from 23.85 to 23.43 that suggests reusing the scene representation to render multiple targets may noticeably improve the training efficiency. Next, we compare the baseline model architecture and an architecture using a separate pose estimation branch and simple token mixing (similar to RUST), which is modified to use a DINOv2 feature extractor and patch-based decoders. This configuration performs comparably to the original RUST \cite{Sajjadi2023} while having significantly better computational speed. LORA based fine-tuning of the feature extractor provides a significant performance boost, suggesting that even a robust foundational model as DINOv2 may need adjustment to extract features for 3D scene synthesis task. Increase of the model size from B to L significantly improves the performance suggesting that scaling up ERUPT model may further improve the performance.

Finally, we examine the case where only 5\% of target images retain ground truth poses during training, where learning is primarily driven by the estimated pose. At inference, meanwhile, we evaluate the performance on synthesis conditioned by the ground truth pose rather than by the estimated pose. We observe only a marginal decrease in performance compared to the baseline, while maintaining the ability to manipulate the camera at inference. UpSRT \cite{Sajjadi2022}, which similar to ERUPT provides camera control at inference and uses unposed input, exhibits a significant performance degradation to 18.64 PSNR when trained on noisy poses \cite{Sajjadi2023}. RUST \cite{Sajjadi2023} trained on unposed images, meanwhile, also reports the ability to manipulate the camera using latent pose instead of ground truth target image, but the corresponding performance is not evaluated.

\begin{table}[htb]
    \centering
    \begin{tabular}{lcccc}
    \toprule
    {Method} & {res} & \multicolumn{2}{c}{\textit{F (ms)}} & \multicolumn{1}{c}{\shortstack{\textit{F} + \textit{B} (ms)}}\\
\cline{3-5}
                   &                    & \textit{s} & \textit{d} & \textit{s + d}\\
    \hline
    ERUPT B & 224 & 6.0 (2.9) & 1.5 & 15\\
    ERUPT L & 224 & 14 (9.4) & 2.3 & 27\\
    SRT & 128 & 3.0 (1.0) & 4.7 & 27(2.6)\\
    RUST \& late SRT & 128 & 5.3 (0.8) & 14 &73(2.3) \\
    SRT & 224 & 11 & 49 & 168 \\
    RUST \& late SRT & 224 & 30 & 118 & OOM \\
    \bottomrule
    \end{tabular}
    \caption{The time of forward and backward pass per scene. \textit{F}, \textit{B}, \textit{s}, \textit{d} stands for forward, backward, scene, and decode, respectively, while res refers to the image size. The time spent by the feature extractor is listed in parentheses.} 
    \label{tab:performance}
\end{table}

\subsection{Computational Performance}

The computational performance of ERUPT is summarized in \cref{tab:performance}. We analyze the time needed to perform forward and backward passes for scene representation generation (‘scene’) and single image generation (‘decode’) on the Ada A6000 GPU. We exclude SD based rendering, which provides high quality output but at about 1 fps for 512x512 resolution. The batch size is selected to maximize VRAM usage and the time measurements are normalized to provide per-scene values.  Mixed precision is enabled in all experiments. We compare our method with the original SRT \cite{Sajjadi2022} and late SRT/RUST \cite{Sajjadi2023} architectures. Since the corresponding papers do not provide the code, we use the implementation from \cite{srtrepo}. The late SRT/RUST setup has a small patch size of 8×8 and 5 scene transformer blocks instead of 10 in the original SRT.

The time spent for scene representation generation is comparable for both models, and the discrepancy is associated with a small patch size used by RUST. Meanwhile, the speed of the ERUPT decoder is an order of magnitude faster than that of RUST, depending on the output resolution. This is a result of computational efficiency gained using our patch-based decoder in comparison to pixel-based decoders used in RUST. The time of the forward and backward pass (corresponding to training speed) of ERUPT-B model at 224-pixel resolution is 5 times shorter than one for RUST at 128-pixel resolution. Finally, the use of frozen DINO-v2 encoder does not introduce a significant overhead during training compared to a simple convolutional extractor in RUST/SRT: 2.9 vs. 2.6 and 2.3 respectively.

The RUST model cannot fit on a single Ada A6000 GPU with 48 GB when using 224-pixel resolution even when using a batch size of 1 because of insufficient VRAM (if all pixels in the image are generated). This is denoted as ‘OOM’ in \cref{tab:performance}. The VRAM required by ERUPT-B is approximately 16 times smaller using a 128-pixel configuration and 64 times smaller using a 224-pixel compared to corresponding RUST model requirements mainly due to ERUPT’s memory friendly patch-based decoder.

\subsection{Real scenes (Mapillary data from MSVS-1M) }

\begin{figure}
\centering
\includegraphics[width=1.0\linewidth]{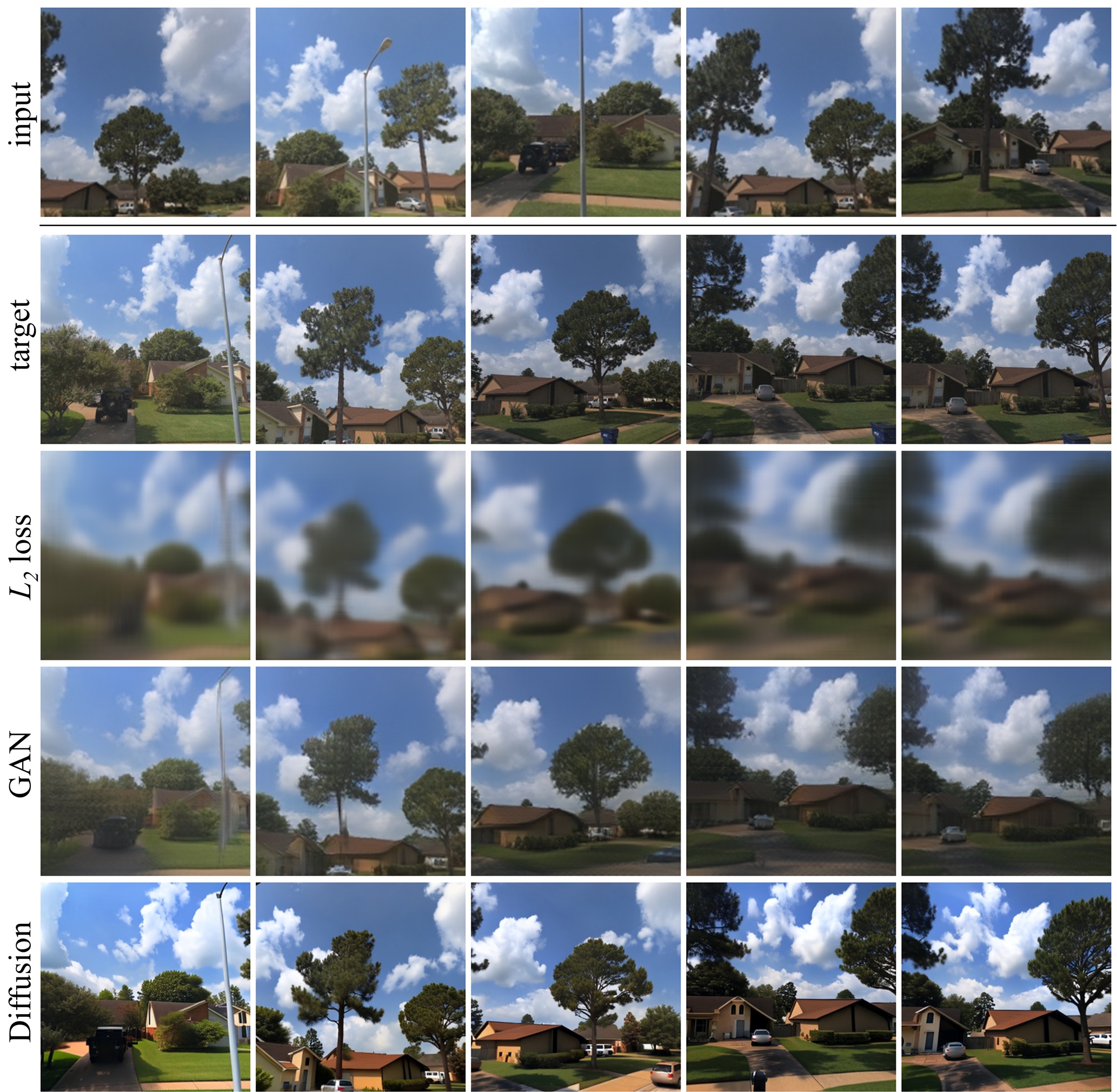}
\caption{\label{fig:EruptResultsMSVS}Qualitative results on MSVS-1M: input scene (5 images), target views (5 images), ERUPT B + LORA, ERUPT B + GAN, ERUPT B + Diffusion.}
\end{figure}

In contrast to MSN, MSVS-1M \cite{msvs1m} is a challenging real-world dataset which captures many of the typical challenges associated with “in-the-wild” data including high object variability, incomplete and discontinuous scenes, transient objects, and camera pose inaccuracy. These present a challenge to view synthesis methods designed strictly for pristine conditions, and better reflect the expected performance of such models in real operating cases. We present this dataset to the community with the hope that it will assist in development of robust view synthesis methods which are suitable for real image datasets. Quantitative results on street-view data from MSVS-1M are provided in \cref{tab:MSVS-1M-result}, and an example of the model output is shown in \cref{fig:EruptResultsMSVS}.

% \begin{table*}[h]
%     \centering
%     \begin{tabular}{lcccccc}
%         \toprule
%         Setup & PSNR↑ & SSIM↑ & LPIPS↓_{Alex} & LPIPS↓_{VGG} & FID↓ \\
%         \midrule
%         ERUPT B + LORA & 20.64 & 0.6572 & 0.8366 & 0.7025 & 442 \\
%         ERUPT B + GAN & 21.07 & 0.6385 & 0.8861 & 0.7266 & 465 \\
%         ERUPT B + SD & 17.10 & 0.5057 & 0.2728 & 0.3714 & 12.6 \\
%         \bottomrule
%     \end{tabular}
%     \caption{Quantitative Results on MSVS-1M dataset}
%     \label{tab:MSVS-1M-result}
% \end{table*}

\begin{table}[h]
    \centering
    \begin{tabular}{lcccccc}
        \toprule
        Method & PSNR↑ & SSIM↑ & LPIPS↓ & FID↓ \\
        \midrule
        ERUPT B+LORA & 20.64 & 0.657 & 0.703 & 442 \\
        ERUPT B+GAN & 21.07 & 0.639 & 0.727 & 465 \\
        ERUPT B+SD & 17.10 & 0.506 & 0.371 & 12.6 \\
        \bottomrule
    \end{tabular}
    \caption{Quantitative results on MSVS-1M dataset.}
    \label{tab:MSVS-1M-result}
\end{table}

As expected, our experiments with ERUPT on this data show that the overall model performance is lower than that reported for MSN dataset. Perceptual quality metrics are especially impacted in the LORA and GAN configurations which may be attributed to blurry images (in case of $L_{2}$ loss) and GAN artifacts. Interestingly, the ERUPT setup using Stable Diffusion (SD) based rendering, shows improved perceptual quality and good alignment to the ground truth.

% As expected, our experiments with ERUPT on this data show that the overall model performance is lower than that reported for MSN dataset. Perceptual quality metrics are especially impacted in the LORA and GAN configurations which may be attributed to blurry images (in case of $L_{2}$ loss) and GAN artifacts. Interestingly, the ERUPT setup using SD rendering shows improved perceptual quality and good alignment to the ground truth.

% \begin{figure}[!ht]
% \centering
% \includegraphics[width=1.0\linewidth]{FiguresAndTables/ERUPT_experiment2.png}
% \caption{\label{fig:EruptExperiment2}}
% \end{figure}

\subsection{Limitations}
One limitation of our approach is its reliance on an $L_{2}$ objective during training. In the case of the MSN dataset, when the inputs typically fully define the scene, and the diversity of objects in the scene is limited, the $L_{2}$ objective enables effective training and provides competitive results. However, if generation of the target image involves out-painting areas not present in the input images or cases where input images have low overlap, the $L_{2}$ objective produces blurry output and may inhibit training from focusing on realistic image generation. Adversarial loss in our GAN-based setup partially addresses this issue, but still suffers from visual artifacts for highly diverse scenes like those in MSVS-1M. Use of diffusion models conditioned on ERUPT for rendering novel views drastically improves the quality of the output especially in the case of MSVS-1M data. However, under conditions of limited input, the diffusion model may produce inconsistent visual features between views. This limitation may be addressed with multi-view diffusion applied in DORSal \cite{jabri2024dorsaldiffusionobjectcentricrepresentations}. Finally, under conditions of limited view our approach with $L_{2}$ objective is likely to under-perform setups trained fully on diffusion objective, \textit{e.g.} recent CAT3d \cite{gao2024cat3d}.

\section{Conclusions}
We present ERUPT, a latent view synthesis model designed for efficient unposed scene representation learning in highly unconstrained environments. Our novel pose-estimation strategy and patch-based encoder-decoder transformer architecture learns to efficiently render novel views, training on both posed and unposed images. We demonstrate an effective method for fully-controllable view synthesis, where target images can be directly queried by specifying camera pose relative to the input images. Our method achieves an order of magnitude improvement of computational efficiency while providing state-of-the-art accuracy in rendered novel views on benchmark datasets. Additionally, we introduce a large-scale, real-world scene reconstruction dataset containing 1M images from 32k sequences. We anticipate that our innovative and computationally efficient approach and open benchmark dataset will enable significant advancements in large-scale latent scene reconstruction.

\section{Acknowledgments}
% Supported by the Intelligence Advanced Research Projects Activity (IARPA) via Department of Interior/ Interior Business Center (DOI/IBC) contract number 140D0423C0075. The U.S. Government is authorized to reproduce and distribute reprints for Governmental purposes notwithstanding any copyright annotation thereon. Disclaimer: The views and conclusions contained herein are those of the authors and should not be interpreted as necessarily representing the official policies or endorsements, either expressed or implied, of IARPA, DOI/IBC, or the U.S. Government.
Supported by the Intelligence Advanced Research Projects Activity (IARPA) via Department of Interior/ Interior Business Center (DOI/IBC) contract number 140D0423C0075. The U.S. Government is authorized to reproduce and distribute reprints for Governmental purposes notwithstanding any copyright annotation thereon. Disclaimer: The views and conclusions contained herein are those of the authors and should not be interpreted as necessarily representing the official policies or endorsements, either expressed or implied, of IARPA, DOI/IBC, or the U.S. Government.

\bibliographystyle{plain}
\bibliography{eruptBibliography}

\clearpage

\begin{figure*}[t!]
\centering
\includegraphics[width=1.0\linewidth]{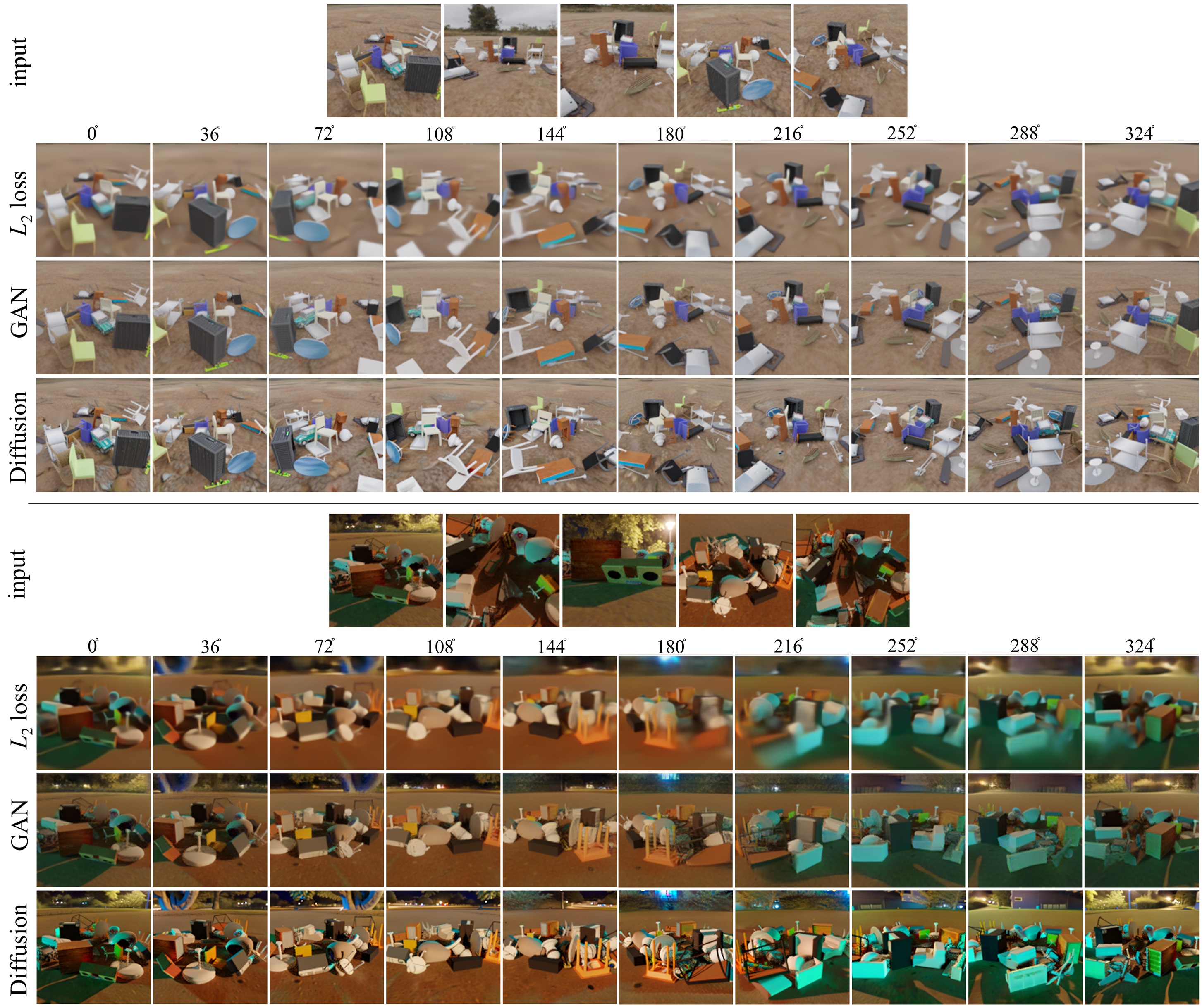}
\caption{\label{fig:360rotate}360-rotation around a scene for cases with all items sufficiently represented in the input images (top) and having missing parts (bottom), which result in blur uncertain output in $L_{2}$ setup.}
\end{figure*}

\section{Supplements}
\subsection{360-rotation illustration}

Figure \ref{fig:360rotate} provides examples of the model output, showcasing a sequence of images generated by 360-rotation around the scene. To give a complete view of ERUPT's scene understanding and synthesis capabilities, we consider two scenarios: one where all items are sufficiently represented in the input views (left), and a more challenging case where input views are far from target views and portions of the scene are not in view of any input imagery (right). When input views are sufficiently dense, ERUPT is able to accurately and consistently synthesize imagery for arbitrary target views after training using a simple $L_{2}$ objective. In the more challenging case where portions of the scene are missing, the $L_{2}$ objective is insufficient resulting in blurry output. Introduction of adversarial loss in our GAN configuration drastically improves the quality of the output eliminating uncertain blurry regions. However, missing information still presents a challenge to reconstruction, resulting in small inconsistencies and GAN artifacts. After finetuning using SD for rendering, challenging target views are reproduced with a high level of detail and fidelity. However, when using fine-tuned Stable Diffusion (SD) for rendering, each novel view is generated independently and uncertain regions may result in inconsistent output between different views.

\subsection{Diffusion Output Consistency}
\begin{figure}[!ht]
\centering
\includegraphics[width=1.0\linewidth]{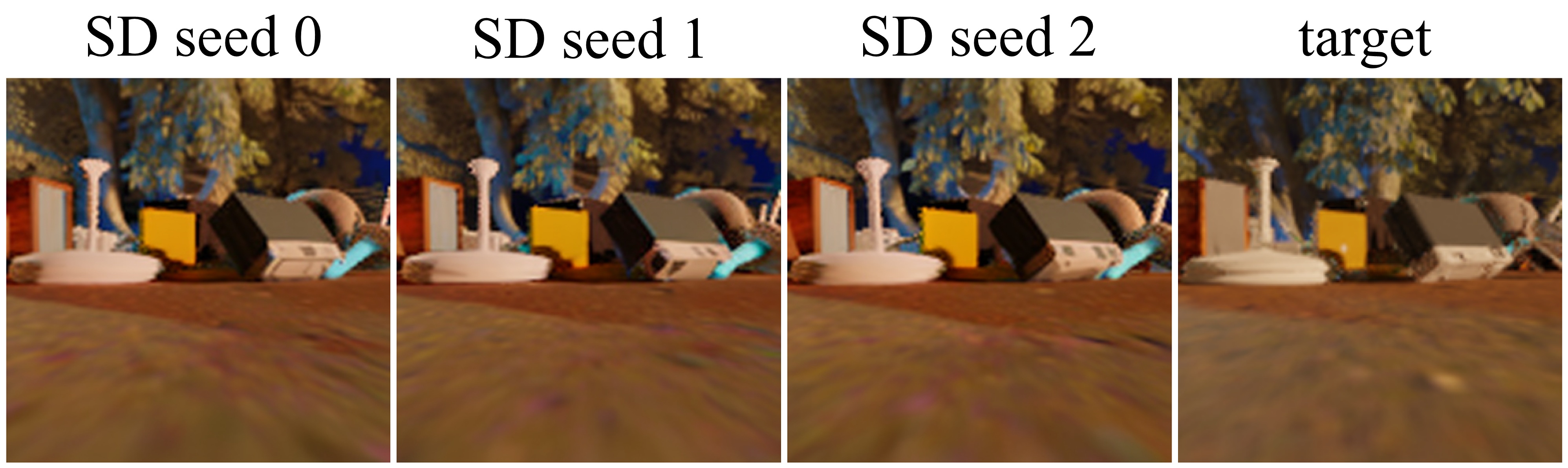}
\caption{\label{fig:EruptExperiment1} Scene consistency: effect of seed on SD output.}
\end{figure}

The output of Diffusion-based models may widely change depending on the provided input noise, which is unfavorable for novel view synthesis. The conditioning provided by ERUPT embedding, however, is sufficient to guide image generation toward semantical and visual coherency. As depicted in Fig. 3 (main text), our approach enables consistent generation of high quality novel views, and in \cref{fig:EruptExperiment1} we provide further illustration of the generation consistency for various input noise (defined by setting random seeds). The images are nearly identical with an insignificant difference in fine details. It should be pointed out, however, that if the input scene has blind spots, the SD output generated for these blind spots may vary.
\subsection{Dynamic Sampling and Scene Creation with Mapillary MSVS-1M Data}
\begin{figure*}[!htp]
\centering
\includegraphics[width=1.0\linewidth]{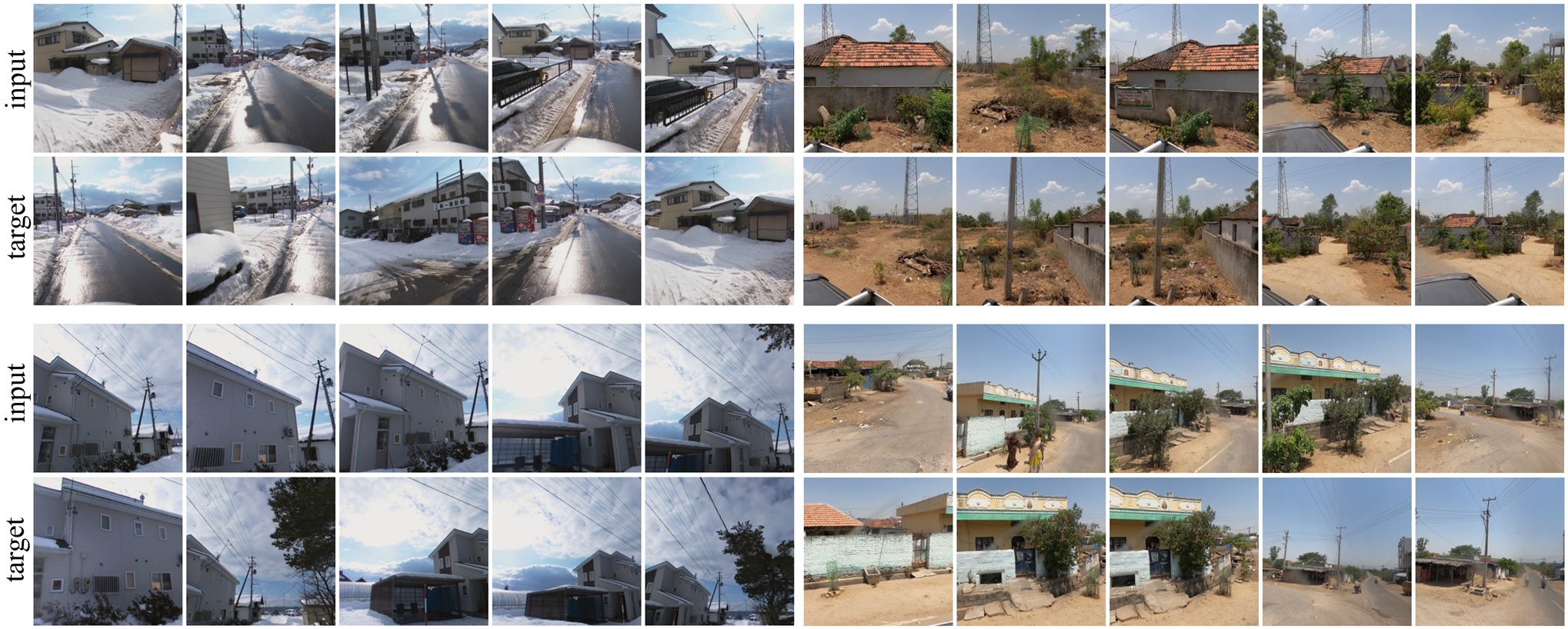}
\caption{\label{fig:MapillarySample} Example scenes created using dynamic sampling of MSVS-1M imagery. Five input views and five target views are shown for each scene. }
\end{figure*}

To create a real dataset of sufficient scale to train our model, we collected a series of panoramic imagery for 10 real-world locations through the Mapillary API (e.g. \hyperlink{https://tiles.mapillary.com}{https://tiles.mapillary.com}). In this section we describe the data collection procedures and present results using ERUPT for novel view rendering on the proposed MSVS-1M dataset. 
The total number of filtered images in the dataset is approximately 1M, which are grouped into 32k continuous sequences (30k train, 944 eval, and 875 test). For training and evaluation we perform dynamic sampling of the original trajectories/panoramas to generate scenes. First, we select a subsequence of the length of 5, and randomly assign it to the input and target locations with repetitions. Next, we select a random reference point at the distance 5-15 of the scene size and sample view directions from the normal distribution with the mean pointing to the reference point and the standard deviation of 0.35 FoV. This strategy is used to make sure that all views cover approximately the same area of the scene but have sufficient diversity. The pitch is selected in the range 0-10 degrees with an additional yaw dependent offset to minimize appearance of the car collecting the data. Finally, we perform Gnomonic projection [20] to generate the corresponding images with FoV of 60 degrees from the panoramic input. The camera parameters are evaluated according to the selected view. At evaluation we draw 48 random scenes from each trajectory. The resulting dataset, MSVS-1M will be released to the public under the CC-BY-SA license.
\cref{fig:MapillarySample} shows several example scenes sampled from the MSVS-1M dataset using the previously described sampling procedure. The MSVS-1M dataset is designed to provide challenging, real-world cases for benchmarking novel view synthesis methods. For example, despite selecting 5 image acquisition locations and pointing the views to the reference point, some target views may include parts of the scene never provided in the input. Additionally, the dataset contains a large variety of locations, and items in contrast to the MSN dataset which includes only ShapeNet items. Another challenge present in the dataset is poor or no overlap between parts of the scene, especially in case of the reference image, which is illustrated at the right bottom part of \cref{fig:MapillarySample}. Under conditions of poor overlap, scene understanding by the model is challenging, and overall, we observe lower performance of the ERUPT model on Mapillary data compared to MSN, however we believe the data serves as a helpful step toward view synthesis on large-scale real world data.

\begin{figure*}[!htp]
\centering
\includegraphics[width=1.0\linewidth]{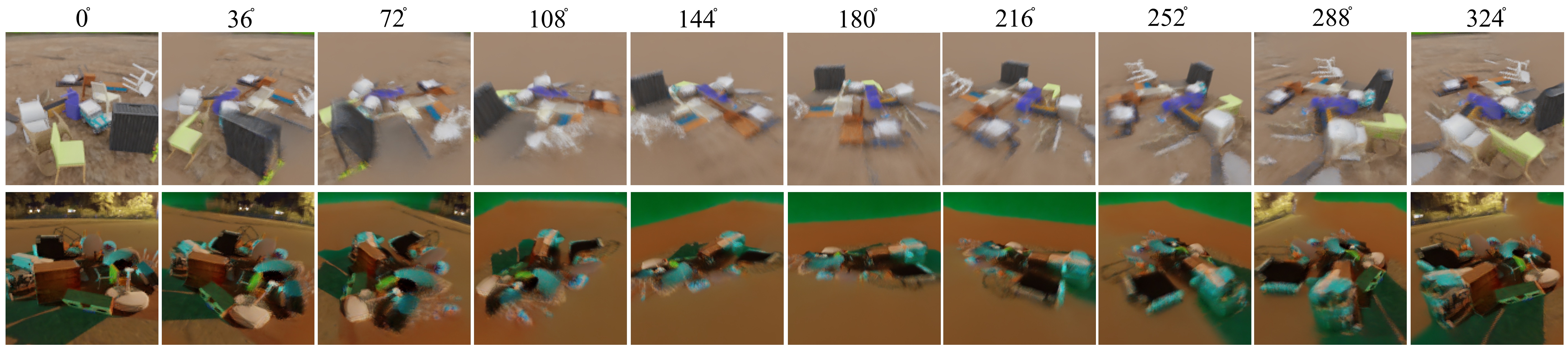}
\caption{\label{fig:zeroNVS} Example of output for scenes from \cref{fig:360rotate} generated with ZeroNVS. Only reference view is used as the input.}
\end{figure*}

\subsection{Loss Functions}
The loss function used for training ERUPT is described by the following:
\begin{equation}
L_{tot}=L_{img}+w_{c}L_{c}+w_{t}L_{t}\label{eq:1}
\end{equation}
where the first term, $L_{img}$, is image $L_{2}$ loss applied between the model output $x^{img}$ and target $y^{img}$ images as shown in Equation \eqref{eq:2}. This term is included to minimize the average error between the target and reconstructed image, and maintain overall scene structure.
\begin{equation}
L_{img}={\left \lVert x^{img}-y^{img} \right \rVert}^{2}\label{eq:2}
\end{equation}
$L_{c}$ is the camera auxiliary loss applied to both the estimated input and target camera parameters, given by \eqref{eq:3}. This loss is included for all experiments except the experiment with 5\% labeled target images, where this loss is omitted. $w_{c}$ is a weighting parameter which controls the contribution of the camera auxiliary loss equal to 1/20. $x^{view}_{k} $,$y^{view}_{k} $ are the $k$-th component of the predicted and ground truth camera basis vectors ($xyz$), respectively, and the corresponding term represents negative cosine loss; while $x^{pos}$, $y^{pos}$  are predicted and ground truth camera positions.
\begin{equation}
L_{c} = \frac{1}{3}\Sigma^{3}_{k=1} (1 - \frac{x^{view}_{k} \cdot y^{view}_{k}}{\left \lVert x^{view}_{k} \right \rVert \left \lVert y^{view}_{k} \right \rVert}) + \frac{1}{20} {\left \lVert x^{pos}-y^{pos} \right \rVert}^{2}\label{eq:3}
\end{equation}
$L_{t}$ is the pair-wise contrastive token loss (see Section 3.1 in the main text) computed for N tokens within each image, as shown in Equation \eqref{eq:4} where $s$=20 and $m$=0.5 are scale and margin, $\theta_{ij}$ is the angle between $i$-th and $j$-th tokens of the considered image, and $L_{t}$ is equal to 1/5. This loss is applied to token decoder output and is included to guide learning the semantics of the scene.

\begin{equation}
L_{t}=\frac{1}{N}\Sigma^{}_{N}log(\frac{e^{s \cdot cos(\theta_{ii+m})}}{e^{s \cdot cos(\theta_{ii+m})} + \Sigma^{}_{N}e^{s \cdot cos(\theta_{ij})}})\label{eq:4}
\end{equation}
  
In the case of our GAN setup, we follow a standard loss formulation with several terms including $L_1$, perceptual and adversarial components written as:
\begin{equation}
L_{tot}^{GAN}=L_{img}^{1}+w_{c}L_{c}+w_{t}L_{t}+L_{p}+L_{g}\label{eq:5}
\end{equation}

where $L_{img}^{1}$ is $L_{1}$ loss applied between input and target images, $L_{p}$ is VGG based perception loss, $L_{g}$ is the GAN loss.

In case of SD finetuning, the ERUPT model is frozen, and we add positional encoding to the output of the token decoder followed by learnable projection to the SD prompt dimension. SD U-net and the projection layers are finetuned with standard epsilon $L_{2}$ diffusion objective. 

\subsection{Comparison to ZeroNVS}
A series of recent studies consider the task of 3d scene generation based on a single input view using a combination of diffusion models and NeRF. \cref{fig:zeroNVS} provides a qualitative comparison of the ZeroNVS output [Sargent \textit{et al.}, ZeroNVS: Zero-Shot 360-Degree View Synthesis from a Single Image] for scenes considered previously with ERUPT, \cref{fig:360rotate}. Only the reference view is used as input to the model. Even if the model is able to reconstruct a few objects in several views, the overall output quality is noticeably lower compared to ERUPT coupled with finetuned SD, which suggests that the use of multiple input views may be crucial for accurate scene reconstruction. In addition, the ZeroNVS runtime is several hours per scene compared to several seconds to ERUPT if SD rendering is used.

\end{document}